\documentclass[sigconf]{acmart}

\usepackage{booktabs} 

\usepackage[T5,T1]{fontenc}
\usepackage{graphicx}
\usepackage{amsmath}
\usepackage{url}
\usepackage{amsmath}
\usepackage{amssymb}

\usepackage{url}      

\usepackage{pstricks}
\usepackage{pst-node} 
\usepackage{pst-tree}
\usepackage{graphicx}

\usepackage[english]{babel}
\usepackage{balance}

\newcommand\tabhead[1]{\small\textbf{#1}}

\begin{document}


\setcopyright{rightsretained}

\copyrightyear{2018}
\acmYear{2018}
\setcopyright{iw3c2w3}
\acmConference[WWW '18 Companion]{The 2018 Web Conference Companion}{April
23--27, 2018}{Lyon, France}
\acmBooktitle{WWW '18 Companion: The 2018 Web Conference Companion, April
23--27, 2018, Lyon, France}
\acmPrice{}
\acmDOI{10.1145/3184558.3191535}
\acmISBN{978-1-4503-5640-4/18/04}

\fancyhead{}

\title{
  A Factoid Question Answering System for Vietnamese
}

\author{Phuong Le-Hong}
\affiliation{Hanoi University of Science, Vietnam}
\affiliation{FPT Technology Research Institute, FPT University}
\email{phuonglh@vnu.edu.vn}
\author{Duc-Thien Bui}
\affiliation{FPT Technology Research Institute}
\affiliation{FPT University, Hanoi, Vietnam}
\email{thienbd@fpt.edu.vn}

\begin{abstract}
  In this paper, we describe the development of an end-to-end factoid
  question answering system for the Vietnamese language. This system
  combines both statistical models and ontology-based methods in a
  chain of processing modules to provide high-quality mappings from
  natural language text to entities. We present the challenges in the
  development of such an intelligent user interface for an isolating
  language like Vietnamese and show that techniques developed for
  inflectional languages cannot be applied ``as is''. Our question
  answering system can answer a wide range of general knowledge
  questions with promising accuracy on a test set.
\end{abstract}



\maketitle

\section{Introduction} 
\label{sec:introduction}

Question Answering (QA) has been an important line of research in
natural language processing in general and human-machine interface
in particular. The ultimate goal of a QA system is to provide a
concise and exact answer to a question asked in a natural
language. For example, the answer to the question ``\textit{Which
  French city has the largest population?}'' should be
``\textit{Paris}''.

Open-domain QA is a challenging task because the research and
validation of a precise answer to a question require a good
understanding of the question itself and of the text containing the
potential answer. Typically we need to carry out both
syntactic and semantic analyses in order to fully understand a
question and pinpoint an answer. This is much more difficult than
the task of common information retrieval, where one only needs to present a
ranked list of documents in response to a question, which can be
efficiently performed by available search engines.

The state-of-the-art techniques in open-domain QA can be classified
into two main categories, namely semantic parsing based techniques and
information retrieval based techniques~\cite{Bordes:2014}. Semantic
parsing systems try to interpret the meaning of a question correctly
by semantic analysis. A correct interpretation converts the
question into an exact database query that returns a correct
answer. On the other hand, information
retrieval based systems first transform a question into a valid query,
then retrieve a set of candidate answers by querying a
corpus and/or a knowledge base, and finally use fine-grained
heuristics to identify the exact answer. 

Although both kinds of system require human expertise to hand-craft linguistic
resources including lexicons, grammars and knowledge bases, the
information retrieval based approach is more suitable to
less-resourced languages since many advanced natural language
processing tools such as syntactic and semantic parsers are not
readily available. Furthermore, as shown in many previous studies on
building QA systems, existing methods developed for well-studied
languages are not easily and conveniently applied or scaled up to
natural languages other than English. 

In this paper we present a QA system for the Vietnamese language
which combines both statistical models and knowledge-based
methods in a chain of processing modules to provide
high-quality mappings from natural language text to entities. We
present the challenges in the development of 
such an intelligent user interface for an isolating language such as
Vietnamese and show that techniques developed for inflectional
languages cannot be applied ``as is''. Our question answering system
can answer a wide range of general knowledge questions with a
promising accuracy on a test set. The system is released as 
open-source software in the hope that it will serve as a baseline for
future developments of question answering systems for Vietnamese.

The remainder of this paper is structured as follows. First, the next section
gives a survey of existing work in the line of this research. Next, we describe
the methodology that we use to develop our QA system. Then, we present our
experiments and evaluation results. Finally, the last section concludes the
paper and suggests some directions for future work.

\section{Related Work}
\label{sec:relatedWork}

There have been some existing studies on building and evaluating QA
system for Vietnamese. In this section, we present a survey of
existing work, compare and hightlight the difference between them and
this work.

Tran~\cite{TranMaiVu:2012} discussed a specific QA system for Vietnamese
person named entity which focuses on only ``\textit{who}'',
``\textit{whom}'' and ``\textit{whose}'' questions. To this end, the
diversity of answerable questions are rather limited. A prior work of
the same research group~\cite{TranMaiVu:2009} presented an
experimental study of a QA system for Vietnamese which utilized a
search engine to search for answers. This system is restricted to
travelling domain and was tested on only a small test set containing
one hundred questions. Duong~\cite{DuongHuuThanh:2014} presented a QA system
for use in Vietnamese legal documents which is able to answer simple
questions about procedures and sanctions in law on business. This
system uses a similarity-based model and the Lucene document search
engine to retrieve candiate documents and extract answers. Compared to
these works, our QA system differs in three aspects. First, it is open
domain which can provide answers to a much wider range of questions
other than a specific domain or person named entity question
types. Second, our system does not use a search engine to retrieve and
rank documents but relies on a large knowledge base. And third, our
system is evaluated on a test set of about ten times larger which
covers a wide variety of questions, resulting in a promising accuracy.

Most recently, Nguyen~\cite{NguyenQuocDat:2016} presented a QA system for
Vietnamese which uses semantic web information to provide answers to
user's queries. Together with a series of previous publications in the
same line of research, this group developed the KbQAS system which is
claimed to be the first knowledge-based QA system for the
language.\footnote{In their work, the term ``knowledge base'' may lead
  to confusion in that it really refers to a set of rules rather than an
  ontology base of entities and relations.}  A key component of their
system is a knowledge acquisition module which utilizes the single
classification ripple down rules method for question analysis. This is
a typical rule-based system. Although their method is able to acquire
rules in a consistent and systematic manner, the knowledge bases are
required to be built from scratch and an adaptation to a new domain or
language still requires time and effort of human expertise. As
reported, this system contains 92 manual rules and was tested in a test set
of 74 Vietnamese questions. In contrast to this work, our system utilizes
both statistical and rule-based approaches, a large ontology base
(DBPedia), and the Cypher query language -- a high-performance
declarative language for query graph database. Our system is also
validated on a much larger test set of diverse questions, totaling
nearly 900 question and answer pairs.

\section{Methodology} 
\label{sec:method}

\subsection{DBPedia and Graph Model}

Our QA system uses an ontology developed by the DBPedia
project~\cite{Lehmann:2015}.\footnote{\url{http://www.dbpedia.org/}}
DBpedia is a crowd-sourced community effort to extract structured
information from Wikipedia and make this information available on the
Web for a wide number of languages, including Vietnamese. The DBPedia
knowledge bases have become an important source of structured
information on the emerging \textit{Web of Data}~\cite{Morsey:2012}.

\begin{figure}[t]
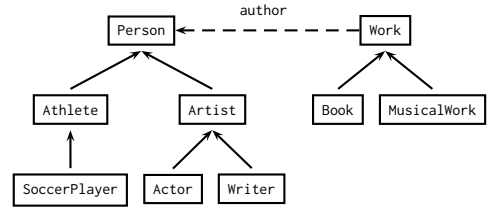

  \centering \psset{arrows=->} {\scriptsize \texttt{
      \begin{tabular}{cc}
        \pstree[nodesep=2pt,levelsep=30pt,radius=2pt,treesep=0.2cm]{\TR[name=A]{\psframebox{Person}}} {
        \pstree{\TR{\psframebox{Athlete}}}{
        \TR{\psframebox{SoccerPlayer}}
        }
        \pstree{\TR{\psframebox{Artist}}}{
        \TR{\psframebox{Actor}}
        \TR{\psframebox{Writer}}
        }
        } 
        &  
          \pstree[nodesep=2pt,levelsep=30pt,radius=2pt,treesep=0.2cm]{\TR[name=B]{\psframebox{Work}}} {
          \TR{\psframebox{Book}}
          \TR{\psframebox{MusicalWork}}
          }
          \ncline[linestyle=dashed]{->}{B}{A}\taput{author}
        \\
      \end{tabular}
    }}
  \caption{An excerpt of the DBPedia ontology} \label{fig:1}
\end{figure}

DBPedia is an ontology according to the definition of
W3C\footnote{\url{http://www.w3.org/standards/semanticweb/}} in that
it defines the terms used to describe and represent an area of
knowledge. Figure~\ref{fig:1} shows an excerpt of the DBPedia
ontology. This ontology says that there is a class called
\texttt{Writer} which is a subclass of \texttt{Artist} which is in
turn a subclass of \texttt{Person}. There is a property relating an
instance of the class \texttt{Work} to an instance of the class
\texttt{Person}. For example, the novel titled ``\textit{Angel and
  Daemon}'' is an instance of class \texttt{Work} and related via
property \texttt{author} to its author ``\textit{Dan Brown}''.

The DBPedia ontology can also be viewed as a property graph model
made up of nodes, relationships and
properties~\cite{Robinson:2015}. Nodes contain properties in the form
of key-value pairs; the keys are strings and the values are arbitrary
data types. Relationships connect and structure nodes. A relationship
always has a direction, a label, a start node and an end node; the
direction and label add semantic clarity to the structuring of
nodes. It is noted that like nodes, relationships can also have
properties which provide not only additional semantics but also
metadata for graph algorithms and help constrain queries at runtime.

We have constructed a graph database from the dump files of the
Vietnamese DBPedia ontology. The total size of these files is about 5
GB. The database is of size 1.5GB, consisting of one million nodes,
2.5 million links and 7.5 million properties.

\subsection{Query Language}

We use Cypher to query the DBPedia ontology. Cypher is an expressive
and compact graph database query language. This language is specific
to Neo4j\footnote{\url{http://www.neo4j.com/}} which is a good and
well-known graph database used by many organizations in production
applications. The major advantages of Cypher are that it is easy to
learn, easy to use and ideal for describing graphs programmatically in
a precise fashion.

It is noted that other graph databases have other means of querying
data. Many graph databases support the RDF query language
SPARQL. However, in building a QA system, we are interested in the
expressive power of a property graph combined with an advantageous
delarative query language. For this reason we chose Cypher to query
the database to find data matching a specific pattern.

\begin{figure}
  \centering \psset{arrows=->,unit=2}
  \begin{pspicture}(0,-0.1)(1,1) \cnodeput(0.5,1){a}{$a$}
    \cnodeput(0,0){b}{$b$} \cnodeput(1,0){c}{$c$}
    \ncline{b}{a}\naput{{\scriptsize\texttt{knows}}}
    \ncline{c}{b}\nbput{{\scriptsize\texttt{knows}}}
    \ncline{a}{c}\naput{{\scriptsize\texttt{knows}}}
  \end{pspicture}
  \caption{A simple graph pattern, expressed using a diagram}
  \label{fig:2}
\end{figure}
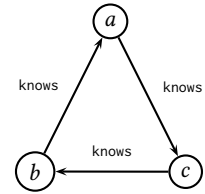

Cypher enables a user (or an application) to ask the database to find
data that matches a specific pattern. Figure~\ref{fig:2} shows an
example of a simple pattern. This pattern describes three mutual
friends.

Like most query languages, Cypher is composed of clauses. The simplest
queries consist of a \texttt{START} clause followed by a
\texttt{MATCH} and a \texttt{RETURN} clause. An example of a Cypher
query that uses these three clauses to find the mutual friends of user
named \texttt{Michael} is:
\begin{verbatim}
START a = node:user(name="Michael")
MATCH (a)-[:knows]->(b)-[:knows]->(c), 
  (a)-[:knows]->(c)
RETURN b, c 
\end{verbatim}
The other clauses that we can use in a Cypher query include
\texttt{WHERE}, \texttt{CREATE} and \texttt{CREATE UNIQUE},
\texttt{DELETE}, \texttt{SET}, \texttt{FOREACH}, \texttt{UNION} and
\texttt{WITH}. These clauses allow for expressive and efficient
querying and updating of the graph database. For details please refer
to the documentation page of the Cypher query
language.\footnote{\url{http://neo4j.com/docs/stable/cypher-query-lang.html}}


In the following subsections we present briefly some important Vietnamese
language processing modules which are integrated in our QA system. These modules
deal with basic processing tasks of Vietnamese including word segmentation,
part-of-speech tagging and question classification. Due to space restriction, we
do not present in this paper the general characteristics of the Vietnamese
language which are discussed in detail in~\cite{Le:2015a}.

\subsection{Word Segmentation}

Word segmentation or tokenization is the problem of dividing a string
of written language into its component words. In English and many
occidental languages, the space is a good approximation of a word
delimiter. However, many languages do not have a trivial word
segmentation process. For example, in Chinese or Japanese, sentences
but not words are delimited; in Thai and Lao, phrases and sentences
but not words are delimited, and as presented in the previous section,
in Vietnamese syllables but not words are delimited. Word
segmentation is in fact a difficult problem for these languages.

In particular, there are two types of ambiguities that we must deal
with in Vietnamese word segmentation. The first ambiguity is called
\textit{overlap ambiguity} where some adjacent syllables can have
different word segmentations and their validity cannot be determined
completely without resorting to a syntactic or semantic resolution of
the entire sentence. For example, the three-syllable phrase
``{\fontencoding{T5}\fontfamily{cmr}\selectfont\textit{thu\d\ocircumflex{}c \dj\d{i}a b\`{a}n}}'' can have two word segmentations, either
``{\fontencoding{T5}\fontfamily{cmr}\selectfont\textit{(thu\d\ocircumflex{}c
    \dj\d{i}a) (b\`{a}n)}}'' or ``{\fontencoding{T5}\fontfamily{cmr}\selectfont\textit{(thu\d\ocircumflex{}c)
  (\dj\d{i}a b\`{a}n)}}'', 
depending on context. A more complicated example is with a
four-syllable phrase ``{\fontencoding{T5}\fontfamily{cmr}\selectfont\textit{t\h\ocircumflex{} h\d\ohorn{}p \acircumflex{}m ti\'\ecircumflex{}t}}'' where all the words
{\fontencoding{T5}\fontfamily{cmr}\selectfont\textit{t\h\ocircumflex{} h\d\ohorn{}p}}, {\fontencoding{T5}\fontfamily{cmr}\selectfont\textit{h\d\ohorn{}p \acircumflex{}m}}, and {\fontencoding{T5}\fontfamily{cmr}\selectfont\textit{\acircumflex{}m ti\'\ecircumflex{}t}} are valid, and
hence all possibly different overlapping word segmentations are
plausible. The second ambiguity is called the \textit{combination
  ambiguity} where two adjacent syllables can either be divided or
combined to make words. For example, two syllables ``\textit{chanh
  chua}'' can form an adjective which means to have a sharp
tongue, or they can form two words \textit{chanh} and \textit{chua}, a
noun phrase which means a sour lemon.

Although Vietnamese word segmentation is difficult, there exists
efficient approaches to solve this problem which have been published
by the Vietnamese language processing community. In this work
we adopt the approach of~\cite{Le:2008a} which is consistent and has
a good accuracy in the range of 96\%--98\% on different test sets.

\subsection{Part-of-Speech Tagging}

Part-of-speech (POS) tagging, also called grammatical tagging or
word-category disambiguation, is the problem of automatically
determining each word in a sentence as corresponding to a particular
part-of-speech such as noun, verb, adjective, adverb, etc. POS tagging is
not an easy problem since many words can represent more than one part
of speech on different occasions.

For well-studied languages like English or certain other occidental
languages, POS tagging is a solved problem with very high
accuracy, about 97.3\%, which is believed to be as high as human
performance~\cite{Manning:2011}. However the accuracy of Vietnamese
POS tagging is much lower than that of English. The combination of
the best machine learning algorithms and the best features in discriminative
sequence models has achieved an accuracy of about
93.5\%~\cite{Le:2010a}. As presented in the previous section, an
important reason for the inferior accuracy of Vietnamese POS
tagging is its inherent difficulty. It is not easy to determine a
clear syntactic function of many Vietnamese words while syntactic
category mutation is a frequent phenomenon. Furthermore, POS tagging
depends heavily on word segmentation, which is a
difficult task as presented in the previous section.


\subsection{Question Classification}

The first step of understanding a question is to perform question
analysis. Question classification is an important task of question
analysis which detects the answer type of the question. Question
classification helps not only to filter out a wide range of candidate
answers but also to determine answer selection strategies. For example,
if one knows that the answer type is \textit{city}, one can restrict
candidate answers as cities instead of consider every noun phrase of a
document providing the answer.

At first glance, one may think that question classification can be
framed as a text classification task. However, there exists
characteristics of question classification that distinguish it from
the common task. Firstly, a question is relatively short and contains
less word-based information than an entire text. Secondly, a short
question needs deeper analysis to reveal its hidden
semantics. Therefore application of text classification algorithms
\textit{per se} to question classification cannot produce good
results. Furthermore natural languages are inherently ambiguous, thus the
question classification is not trivial, especially for \textit{what}
and \textit{which} type questions. For example ``\textit{What is the
  capital of France?}''  is of location (city) type, while
``\textit{What is the Internet of things?}'' is of definition
type. Consider also these examples: \textit{(1) What tourist attractions are
  there in Reims? (2) What do most tourists visit in Reims? (3) What
  are the names of the tourist attractions in Reims? (4) What attracts
  tourists to Reims? (5) What is worth seeing in
  Reims?}~\cite{Li:2006}; all these questions are of the same answer
type: location. Different wording and syntactic structures classification 
difficult ~\cite{Huang:2008}.

With the increasing popularity of statistical approaches to natural
language processing in general and to question classification in
particular, recent years have seen many machine learning approaches
which have been applied to the problem of question classification. The main
advantage of machine learning approaches is that one can learn a
statistical model using useful features extracted from a sufficiently
large set of labeled questions and then use it to automatically
classify new questions.

We use the method proposed by~\cite{Le:2014} in our question
classification module. In contrast to many existing approaches for
question classification which make use of very rich feature spaces or
hand-crafted rules, this method proposes a compact yet effective
feature set. In particular it uses typed dependencies as semantic
features. It has been shown that by integrating only two simple
dependencies of types nominal subject and prepositional object, one can
improve the accuracy of question classification by over 8.0\% using
common statistical classifiers over two benchmark datasets, the UIUC
dataset for English and a recently introduced FPT question dataset for
Vietnamese. With unigram feature and typed dependency feature, one can
obtain accuracy of 87.6\% and 80.5\% using maximum entropy classification
for the UIUC and FPT question dataset respectively. It is worth noting that
the best question classification accuracy on the UIUC dataset is
89.00\% by~\cite{Huang:2008}, where important features like head words
and their hypernyms are included. Such semantic features are not
readily available for less-resourced languages such as Vietnamese,
where a WordNet is still in its first stage of construction.

\subsection{Textual Question to Cypher Query Transformation}

An important module of our knowledge-based QA system is the module
that transforms textual questions in Vietnamese into equivalent Cypher
queries. The queries are then executed to search for answers to the
questions. This section describes the main processing steps of this
module.

First, a textual question is processed by the NLP chain presented
above, ranging from word segmentation to part-of-speech tagging and
question classification. For example a question such as
``{\fontencoding{T5}\fontfamily{cmr}\selectfont\textit{Th\`{a}nh vi\ecircumflex{}n ch\h{u} ch\'\ocircumflex{}t c\h{u}a t\d\acircumflex{}p \dj{}o\`{a}n FPT l\`{a} nh\~\uhorn{}ng ai?''}} (``Who
are the most important people of FPT Corporation?'') will be analysed
as follows:
\begin{description}
\item[Word Segmentation:] The output is a sequence of words:
  {\fontencoding{T5}\fontfamily{cmr}\selectfont\textit{[Th\`{a}nh\_vi\ecircumflex{}n, ch\h{u}\_ch\'\ocircumflex{}t, c\h{u}a, t\d\acircumflex{}p\_\dj{}o\`{a}n, FPT,
    l\`{a}\_nh\~\uhorn{}ng\_ai]}}. Here the underscore character is used to connect
  the syllables of a word and words are separated by commas.
\item[Part-of-speech Tagging:] The output is a sequence of tagged
  words: {\fontencoding{T5}\fontfamily{cmr}\selectfont\textit{[Th\`{a}nh\_vi\ecircumflex{}n/\texttt{N} , ch\h{u}\_ch\'\ocircumflex{}t/\texttt{N},
    c\h{u}a/\texttt{E}, t\d\acircumflex{}p\_\dj{}o\`{a}n/\texttt{N}, FPT/\texttt{Np},
    l\`{a}\_nh\~\uhorn{}ng\_ai?/\texttt{QW}]}}. In this step, each word of the
  question is tagged as a part-of-speech, where \texttt{N} denotes a
  common noun, \texttt{Np} denotes a proper noun, \texttt{E} denotes a
  preposition and so on.
\item[Key Word Extraction:] In this step stop words or unimportant
  words are stripped out of the question. Only key words are
  retained. In the example above, the word {\fontencoding{T5}\fontfamily{cmr}\selectfont\textit{c\h{u}a/\texttt{E}}} is
  removed.
\item[Question Classification:] This step determines the answer type
  of the question, that is the information type we need to find. In
  this example the answer type is \texttt{HUM} (human) since the
  question asks for a person (or a group of people). Some other answer
  types are \texttt{NUM} (number), \texttt{DTIME} (datetime),
  \texttt{YESNO} (yes/no), etc. Details of the question types,
  statistical models and classification techniques are presented
  in~\cite{Le:2014}.
\item[Entity Construction:] Since we are querying a graph model which
  is made up of entities (nodes), relationships and properties, we
  need to construct a set of entities, relationships and properties
  which are implied in the query at hand. This step is crucial in
  building a good corresponding Cypher query for a textual
  question. In essence:
  \begin{itemize}
  \item The proper nouns correspond to the names of the nodes in the
    graph database. They are recognized by using their part-of-speech
    tag \texttt{Np}.
  \item The remaining words are classified as either properties or
    relationships, depending on their probabilities on a datasets, using
    a built-in dictionary.
  \end{itemize}
  More specifically, our approach combines a rule-based extractor and
  a statistical-based classifier to perform entity construction. A
  rule-based extractor is used to extract named entities such as persons,
  organizations or locations by relying on the output of a
  part-of-speech tagger. A logistic regression model is used to
  predict the likelihood of being a property or a relationship for
  each remaining keyword in the query. 
  
  To continue with the example above, this step determines \texttt{FPT}
  as entity, and {\fontencoding{T5}\fontfamily{cmr}\selectfont\texttt{th\`{a}nhVi\ecircumflex{}nCh\h{u}Ch\'\ocircumflex{}t}} as relationship, and
  there is no property for this question.
\item[Cypher Query Construction:] In this last step we first build a
  syntactic tree representing a cypher query which corresponds to the
  textual input question. The nodes in the syntactic tree correspond
  to either the Cypher clauses or the operators (``\texttt{=}'',
  ``\texttt{>}'', ``\texttt{<}'', \dots), starting from the root node
  whose name is \texttt{START}. The leaf nodes of the tree correspond
  to key words or values. Figure~\ref{fig:3} shows an example of a 
  syntactic tree.
  \begin{figure}
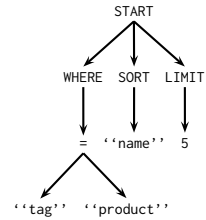

    \centering \psset{arrows=<-}
    {\scriptsize \texttt{
        \pstree[nodesep=2pt,levelsep=25pt,radius=2pt,treesep=0.2cm]{\TR{START}}
        { \pstree{\TR{WHERE}}{
            \pstree{\TR{=}}{
              \TR{``tag''}
              \TR{``product''} } }
          \pstree{\TR{SORT}}{ \TR{``name''} }
          \pstree{\TR{LIMIT}}{ \TR{5} } } } }
    \caption{An example of a Cypher syntactic tree} \label{fig:3}
  \end{figure}
  Once the syntactic tree has been built, we search for appropriate
  replacements of leaf nodes with the elements determined in the
  previous step. Since there may be multiple plausible replacements for
  leaf nodes, a syntactic tree may generate multiple Cypher
  queries. This is expected because in the Vietnamese
  DBPedia graph database a key word can either be a link of a node or
  be a property of another node. A Cypher query for the original
  question is
{\fontencoding{T5}\fontfamily{cmr}\selectfont

  START x=node:DBPediaIndex(key="FPT")
  
  RETURN DISTINCT x.{\fontencoding{T5}\fontfamily{cmr}\selectfont
  th\`{a}nhVi\ecircumflex{}nCh\h{u}Ch\'\ocircumflex{}t}
}
\end{description}

\begin{table*}[t]
  \centering
  \caption{Some question types and examples}
  \label{tab:1}
  {\fontencoding{T5}\fontfamily{cmr}\selectfont
  {\scriptsize
  \begin{tabular}{|p{5cm}|p{5.5cm}|p{6cm}|}
    \hline
    \tabhead{Question Type} & \tabhead{Example} & \tabhead{English Translation}\\
    \hline
    $E$ l\`{a} [ai | g\`{i} | \h\ohorn{} \dj{}\acircumflex{}u | \dots]? & Nguy\~\ecircumflex{}n T\'\acircumflex{}n D\~{u}ng l\`{a} ai? & Who is
    Nguyen Tan Dung?\\ 
    & Facebook l\`{a} g\`{i}? & What is Facebook?\\ \hline 
    $p$ c\h{u}a $E$ [nh\uhorn{} th\'\ecircumflex{} n\`{a}o | b\`\abreve{}ng bao nhi\ecircumflex{}u | \dots]?  
    & D\acircumflex{}n s\'\ocircumflex{} c\h{u}a H\`{a} N\d\ocircumflex{}i b\`\abreve{}ng bao nhi\ecircumflex{}u? &What is the population of Hanoi?\\ \hline 
    $r$ c\h{u}a $E$ l\`{a} g\`{i}? & Th\h{u} \dj{}\ocircumflex{} c\h{u}a Th\'{a}i Lan l\`{a} g\`{i}? &What is the
                         capital of Thailand?\\ \hline 
    $r$ c\h{u}a $E$ [nh\uhorn{} th\'\ecircumflex{} n\`{a}o | b\`\abreve{}ng bao nhi\ecircumflex{}u | \dots ]? & V\d\ohorn{} c\h{u}a th\h{u}
                                                           t\uhorn{}\'\ohorn{}ng
                                                           Nguy\~\ecircumflex{}n T\'\acircumflex{}n
                                                           D\~{u}ng l\`{a} ai? &
                                                           Who is the
    wife of Prime Minister Nguyen Tan Dung?\\ 
    & Ch\h{u} t\d{i}ch HĐQT t\d\acircumflex{}p \dj{}o\`{a}n FPT l\`{a} ai? & Who is the chairman of FPT Corp.? \\ \hline 
    $p$ c\h{u}a $r$ c\h{u}a $E$ l\`{a} g\`{i}? & T\ecircumflex{}n c\h{u}a v\d\ohorn{} vua Tr\`\acircumflex{}n Th\'{a}i T\ocircumflex{}ng l\`{a} g\`{i}? &
                                 What is the name of King Tran Thai
    Tong's wife?\\ 
    & N\ohorn{}i sinh c\h{u}a ch\h{u} t\d{i}ch UBND TP. H\`{a} N\d\ocircumflex{}i \h\ohorn{} \dj{}\acircumflex{}u? & What is the place
    of birth of the chairman of Hanoi People's Committee?\\
    \hline
    $E_1$ v\`{a} $E_2$ c\'{o} $r$ l\`{a} g\`{i}? & Vi\d\ecircumflex{}t Nam v\`{a} Th\'{a}i Lan c\'{o} th\h{u} \dj{}\ocircumflex{} l\`{a}
                                   g\`{i}? & What are the capitals of
                                   Vietnam and Thailand?\\ 
    \hline
  \end{tabular}}}
\end{table*}

As another complete example, consider the following question:
``{\fontencoding{T5}\fontfamily{cmr}\selectfont\textit{D\acircumflex{}n s\'\ocircumflex{} v\`{a} di\d\ecircumflex{}n t\'{i}ch c\h{u}a H\`{a} N\d\ocircumflex{}i l\`{a} bao nhi\ecircumflex{}u?}}'' (What is the
population and area of Hanoi?). This question is analysed by the
processing chain above, where the intermediate results and the final 
Cypher query are as follows:
\begin{enumerate}
\item Word Segmentation and Part-of-Speech Tagging:

  {\fontencoding{T5}\fontfamily{cmr}\selectfont\textit{[D\acircumflex{}n\_s\'\ocircumflex{}/\texttt{N} v\`{a}/\texttt{E} di\d\ecircumflex{}n\_t\'{i}ch/\texttt{N}
    c\h{u}a/\texttt{A} H\`{a}\_N\d\ocircumflex{}i/\texttt{Np} \newline
      l\`{a}\_bao\_nhi\ecircumflex{}u?/QW}]}
\item Key Word Extraction: 

  {\fontencoding{T5}\fontfamily{cmr}\selectfont\textit{[D\acircumflex{}n\_s\'\ocircumflex{}/\texttt{N},
    di\d\ecircumflex{}n\_t\'{i}ch/\texttt{N}, H\`{a}\_N\d\ocircumflex{}i/\texttt{Np}]}}
\item Question Classification: The answer type is \texttt{NUM}
\item Entity Construction: Properties = {\fontencoding{T5}\fontfamily{cmr}\selectfont\{D\acircumflex{}n\_s\'\ocircumflex{}, Di\d\ecircumflex{}n\_t\'{i}ch\}; Named
  Entity = \{H\`{a}\_N\d\ocircumflex{}i\}}
\item Cypher Query Construction:
  {\fontencoding{T5}\fontfamily{cmr}\selectfont
    
  START n=node:DBPedia(key="H\`{a}\_N\d\ocircumflex{}i") 

  RETURN n.d\acircumflex{}nS\'\ocircumflex{}, n.di\d\ecircumflex{}nT\'{i}ch
}
\end{enumerate}
In the next section, we present experimental results of our QA
system and discussion.

\section{Experiments}
\label{sec:exp}

Our aim is to build a QA system which is able to answer Vietnamese
factoid questions on a broad range of topics from the DBPedia ontology
with high accuracy. We have developed an algorithm to transform
different questions to corresponding Cypher queries following the
methodology described above. The system can answer a wide variety of
questions of diverse types, which are shown in the
Table~\ref{tab:1}. 

In this table $E, p$ and $r$ represents an entity, a property and a
relationship respectively, and the vertical character '|' is used to
represent alternative choices. The last row of the table shows a
complicated question type where we seek the same relationship of two
different entities (here, the capital). It is also extended further to
account for more complicated questions where a user wants to seek for
some comparative information, such as in the following example
question where the area and population of two different Vietnamese
provinces are queried:
\begin{center}
  {\fontencoding{T5}\fontfamily{cmr}\selectfont``\textit{Di\d\ecircumflex{}n t\'{i}ch v\`{a} d\acircumflex{}n s\'\ocircumflex{} c\h{u}a H\`{a} N\d\ocircumflex{}i v\`{a} Th\'{a}i B\`{i}nh b\`\abreve{}ng bao
    nhi\ecircumflex{}u?}''} (What are the area and population of Hanoi and Thaibinh?)
\end{center}

It is worth noting that the system can effectively deal with different
variants of the same question since different syntactically correct word
orders are identified and analysed. For example, to query the
population of Hanoi, one can use either of the two following
paraphrases:
\begin{center}
  {\fontencoding{T5}\fontfamily{cmr}\selectfont
\begin{tabular}{l}
\textit{D\acircumflex{}n s\'\ocircumflex{} c\h{u}a H\`{a} N\d\ocircumflex{}i l\`{a} bao nhi\ecircumflex{}u?}\\
\textit{H\`{a} N\d\ocircumflex{}i c\'{o} d\acircumflex{}n s\'\ocircumflex{} b\`\abreve{}ng bao nhi\ecircumflex{}u?}\\
(What is the population of Hanoi?)
\end{tabular}}
\end{center}
Or to ask for the country whose capital is Bangkok, a Vietnamese
speaker can use either of the two following choices: 
\begin{center}
  {\fontencoding{T5}\fontfamily{cmr}\selectfont
\begin{tabular}{cc}
\textit{Bangkok l\`{a} th\h{u} \dj{}\ocircumflex{} c\h{u}a n\uhorn{}\'\ohorn{}c n\`{a}o?}\\
\textit{N\uhorn{}\'\ohorn{}c n\`{a}o c\'{o} th\h{u} \dj{}\ocircumflex{} l\`{a} Bangkok?}\\
(What country's capital is Bangkok?)
\end{tabular}
}
\end{center}

To evaluate the performance of the system, we have manually built a dataset of
879 question-answer pairs about person, location and other facts 
where the answers can be found in the Vietnamese
Wikipedia.\footnote{Currently, the Vietnamese Wikipedia
contains about $1,140,000$ articles according to \url{https://stats.wikimedia.org/EN/}.} 
To understand the performance further, in addition to the accuracy of the final answer, 
the accuracy of query transformation is also evaluated. The accuracy of our system 
is shown in Table~\ref{tab:2}.

\begin{table}[t]
  \caption{Accuracy of the system} \label{tab:2}
  \centering
  \begin{tabular}{|l|l|}
    \hline
    Accuracy of the QA system&76.90\%\\
    \hline
    Accuracy of the query construction module&97.50\%\\
    \hline
  \end{tabular}
\end{table}

The system is able to give correct answers for $76.70\%$ of the questions in the test set.
If the system does not find an answer to a question, it is counted as an incorrect result 
for that question. 

The current test set contains the following types of questions of
different difficulty levels:
\begin{itemize}
\item Questions about an entity of the form ``\textit{Who/What/\dots
    is E}''. For example, ``\textit{Who is Barack Obama?}'' or "\textit{Where
  is Paris?}'' (in Vietnamese language).
\item Questions about a feature or property of an entity, for example
  ``\textit{Who is the spouse of Barack Obama?}'', ``\textit{What is
    the population density of Hanoi?}'', or a trickier question such
  as \textit{``What is the population of the capital of Argentina?}.
\item Questions about the same relationship of two different entities,
  for example ``\textit{What are the capitals of France and
    Germany?}. Here, France and Germany are two entities and the same
  relationship is the capital. The correct answer for this question
  should be ``Paris'' and ``Berlin''.
\end{itemize}
Table~\ref{tab:1} gives some more examples of these types of questions
along with their English translation. 

Our manual test set also contains the correct Cypher query for each
question so that the automatic query construction module can also be
evaluated. Table~\ref{tab:3} shows two samples in our test set.

\begin{table*}[t]
  \centering
  \caption{Some samples in our test set}
  \label{tab:3}
  \includegraphics[scale=0.6]{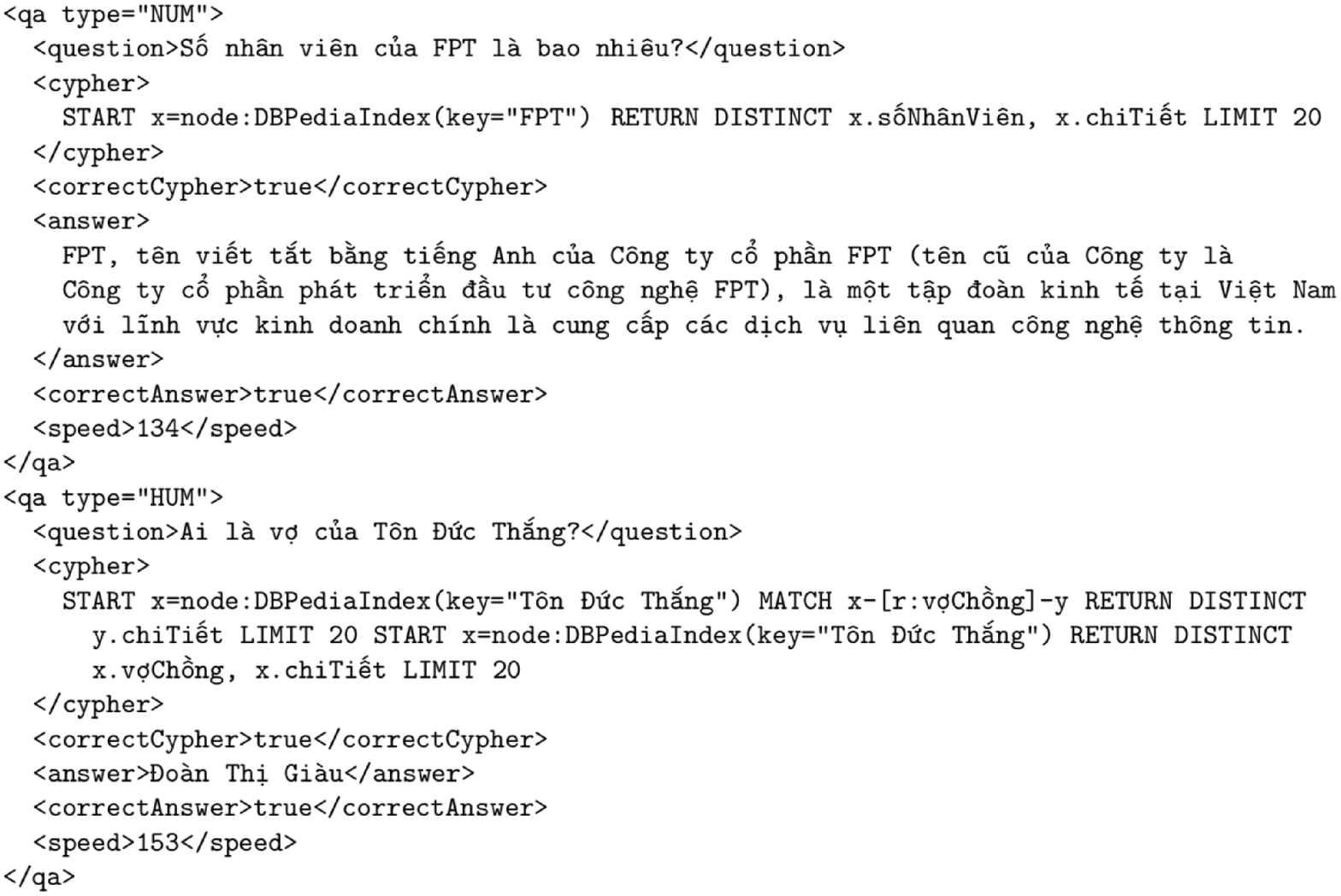}
\end{table*}

Our QA system is designed to provide a short answer to factoid
questions. However by querying the DBPedia graph database built on top
of the Wikipedia it can also answer many questions in great
details in the sense that it can also show the answer text for the
question whenever it is available. For example, the following snippet shows the answer to the
question {\fontencoding{T5}\fontfamily{cmr}\selectfont``\textit{T\'{a}c gi\h{a} c\h{u}a Truy\d\ecircumflex{}n Ki\`\ecircumflex{}u l\`{a} ai?}}'' (Who is the
author of The Tale of Kieu?). Once the system finds the short answer
``{\fontencoding{T5}\fontfamily{cmr}\selectfont\textit{Nguy\~\ecircumflex{}n Du}}'' for this question, it can retrieve and show the summary
of the corresponding entry in the Wikipedia, giving the following result:

\noindent
{\fontencoding{T5}\fontfamily{cmr}\selectfont
\textit{ 
<question>T\'{a}c gi\h{a} c\h{u}a Truy\d\ecircumflex{}n Ki\`\ecircumflex{}u l\`{a} ai?</question>\\
<answer>Nguy\~\ecircumflex{}n Du t\ecircumflex{}n ch\~\uhorn{} T\'\ocircumflex{} Nh\uhorn{}, hi\d\ecircumflex{}u Thanh Hi\ecircumflex{}n, bi\d\ecircumflex{}t hi\d\ecircumflex{}u 
H\`\ocircumflex{}ng S\ohorn{}n l\d{a}p h\d\ocircumflex{}, l\`{a} m\d\ocircumflex{}t nh\`{a} th\ohorn{} n\h\ocircumflex{}i ti\'\ecircumflex{}ng th\`\ohorn{}i L\ecircumflex{} m\d{a}t, Nguy\~\ecircumflex{}n s\ohorn{} \h\ohorn{} Vi\d\ecircumflex{}t Nam. 
\Ocircumflex{}ng l\`{a} m\d\ocircumflex{}t nh\`{a} th\ohorn{} l\'\ohorn{}n c\h{u}a Vi\d\ecircumflex{}t Nam, \dj{}\uhorn{}\d\ohorn{}c ng\uhorn{}\`\ohorn{}i
Vi\d\ecircumflex{}t k\'{i}nh tr\d{o}ng g\d{o}i 
\ocircumflex{}ng l\`{a} "\DJ\d{a}i thi h\`{a}o d\acircumflex{}n t\d\ocircumflex{}c". N\abreve{}m 1965, Nguy\~\ecircumflex{}n Du \dj{}\uhorn{}\d\ohorn{}c H\d\ocircumflex{}i \dj{}\`\ocircumflex{}ng
 h\`{o}a b\`{i}nh 
 th\'\ecircumflex{} gi\'\ohorn{}i c\ocircumflex{}ng nh\d\acircumflex{}n l\`{a} danh nh\acircumflex{}n v\abreve{}n h\'{o}a th\'\ecircumflex{} gi\'\ohorn{}i v\`{a} ra quy\'\ecircumflex{}t \dj{}\d{i}nh 
k\h{y} ni\d\ecircumflex{}m tr\d{o}ng th\h\ecircumflex{} nh\acircumflex{}n d\d{i}p
 200 n\abreve{}m n\abreve{}m sinh c\h{u}a \ocircumflex{}ng.
</answer>
}
}

Our QA system has a good speed in that it can answer a
question in average $0.04$ second on a personal computer. Our system
will be released as an open-source
project and freely available for research purpose. We believe that
our system will be useful for the Vietnamese language 
processing community. At the moment, our demo system is available for
testing at \url{http://124.158.5.68:8080/wiki-qa/}.

\section{Conclusion}
\label{sec:conclusion}

This paper presented the development of an open-domain question
answering system for the Vietnamese language. The system combines both
statistical models and ontology-based methods in a chain of processing
modules to provide high-quality mappings from natural language text to
entities. It can answer a wide range of general knowledge questions
with promising accuracy on a test set. It is released as an
open-source software project in the hope that it will serve as a
baseline for future development of question answer systems for
Vietnamese.\footnote{The temporary demo link of our system is at
  \url{http://124.158.5.68:8080/wiki-qa/}}.

With the rise of available large scale structured knowledge bases, we
think that the most promising approach to open-domain question
answering is the ability to query efficiently such databases in
natural languages. In this work we concentrated on exploiting DBPedia,
a freely available database of facts which are extracted from the
Wikipedia. Nevertheless there exists other good knowledge bases such
as Freebase~\cite{Bollacker:2008}, an open shared database of the
world's knowledge, which has been shown to be very useful for many
applications including question answering. We plan to investigate how
we can use the Vietnamese section of this knowledge base in our system
in a future work. We also plan to perform some comparisons with other
approaches that can find answers directly from Wikipedia texts to show
the benefit of quering an ontology.

Current good question answering systems make use of additional natural
language processing modules such as dependency parsing or semantic
role labelling~\cite{Cui:2005}. We would like to improve further the
performance of our system by integrating recently available dependency
parser~\cite{Le:2015c}, semantic role labeller~\cite{Pham:2015} and
named entity recognizer for
Vietnamese~\cite{Le:2016}. 

Finally recent works on open-domain question
answering~\cite{Bordes:2014,Bordes:2014b} have shown the efficiency of
embedding models, which learn low dimensional vector representations
of words and knowledge bases constituents to achieve better
accuracy. How these models can be used to improve our current system
is another interesting line of research that we would like to research
in a future work, following some recent results~\cite{Hoang:2017}.


\begin{acks}
This research is partly funded by FPT Technology Innovation, FPT
Corporation. The first author is partly funded by the
Vietnam National University, Hanoi (VNU) under project number
QG.15.04. We are grateful to our anonymous reviewers for their
helpful comments which helped us improve the quality of the article.
\end{acks}

\bibliographystyle{ACM-Reference-Format}
\balance
\bibliography{bibliography}

\end{document}